\documentclass{article}

\usepackage{arxiv}
\usepackage[utf8]{inputenc} 
\usepackage[T1]{fontenc}    
\usepackage{hyperref}       
\usepackage{url}            
\usepackage{booktabs}       
\usepackage{amsfonts}       
\usepackage{nicefrac}       
\usepackage{microtype}      
\usepackage{lipsum}		
\usepackage{tabularx} 
\newcolumntype{Y}{>{\raggedright\arraybackslash}X} 
\usepackage{graphicx}
\usepackage{natbib}
\usepackage{doi}
\usepackage{amsmath}
\usepackage{amssymb}
\usepackage{amsfonts}

\providecommand{\Description}[1]{}

\title{Cascading Multi-Agent Anomaly Detection in Surveillance Systems via Vision-Language Models and Embedding-Based Classification}

\author{
Tayyab Rehman\\
Department of Engineering,\\
Computer Science and Mathematics (DISIM),\\
University of L'Aquila\\
L'Aquila, Italy\\
\texttt{tayyab.rehman@graduate.univaq.it}
\And
Giovanni De~Gasperis\\
Department of Engineering,\\
Computer Science and Mathematics (DISIM),\\
University of L'Aquila\\
L'Aquila, Italy\\
\texttt{giovanni.degasperis@univaq.it}
\And
Aly Shmahell\\
Department of Engineering, \\ Computer Science and Mathematics (DISIM),\\ University of L’Aquila, L’Aquila, Italy\\
SPEE S.R.L, L’Aquila, Italy\\
\texttt{aly.shmahell@student.univaq.it, aly.shmahell@spee.it}
}

\renewcommand{\shorttitle}{\textit{arXiv}}

\begin{document}
\maketitle

\begin{abstract}
Intelligent anomaly detection in dynamic visual environments requires reconciling real-time performance with semantic interpretability. Conventional approaches address only fragments of this challenge: reconstruction-based models capture low-level deviations without contextual reasoning, object detectors provide speed but limited semantics, and large vision--language systems deliver interpretability at prohibitive computational cost. This work introduces a cascading multi-agent framework that unifies these complementary paradigms into a coherent and interpretable architecture. Early modules perform reconstruction-gated filtering and object-level assessment, while higher-level reasoning agents are selectively invoked to interpret semantically ambiguous events. The system employs adaptive escalation thresholds and a publish--subscribe communication backbone, enabling asynchronous coordination and scalable deployment across heterogeneous hardware.  Extensive evaluation on large-scale monitoring data demonstrates that the proposed cascade achieves a threefold reduction in latency compared to direct vision--language inference, while maintaining high perceptual fidelity (PSNR = 38.3 dB, SSIM = 0.965) and consistent semantic labeling. The framework advances beyond conventional detection pipelines by combining early-exit efficiency, adaptive multi-agent reasoning, and explainable anomaly attribution---establishing a reproducible and energy-efficient foundation for scalable intelligent visual monitoring.
\end{abstract}


\section{Introduction}
Anomaly detection in surveillance video has become an indispensable component of intelligent monitoring systems, as large-scale camera deployments generate continuous streams that cannot be reviewed manually. Effective systems must satisfy two conflicting requirements: real-time responsiveness, to react to critical events with minimal latency, and semantic interpretability, to ensure that alerts correspond to meaningful anomalies~\cite{li2024llava}. Conventional methods achieve only partial success. Object-detection pipelines, including recent high-speed variants of the YOLO family, provide rapid recognition but are limited to predefined categories and fail to generalize to complex or unseen events ~\cite{wang2024yolov9}. Reconstruction-based methods, such as convolutional autoencoders, detect deviations from learned standard patterns without requiring explicit labels, but are often unstable under illumination changes, camera noise, or dynamic backgrounds ~\cite{liu2025networking}.

Recent surveys emphasize the potential of vision-language models (VLMs) for video anomaly detection, highlighting their ability to generate natural-language descriptions and capture contextual semantics ~\cite{neloy2024comprehensive}. Emerging frameworks, such as LLaVA-NeXT-Interleave, extend these capabilities to multi-image and video scenarios, while weakly supervised methods, like VadCLIP, demonstrate how textual prototypes can align with video features to detect anomalies without exhaustive annotations ~\cite{wu2024deep}. Further advances in unsupervised VLM-assisted anomaly detection explore promising applications in open-world surveillance contexts. Nevertheless, the high inference latency and computational footprint of VLMs constrain their direct use in real-time deployments. The literature increasingly points toward cascading architectures as a viable solution ~\cite{wu2024vadclip}. Lightweight detectors handle routine inputs at the edge, while only ambiguous or complex cases escalate to computationally intensive reasoning modules. This mirrors trends in networking and multi-agent surveillance, where event-driven responsiveness is complemented by background monitoring and infrastructure health checks. Such designs not only improve efficiency but also align with requirements for privacy preservation and scalability in distributed multi-camera systems ~\cite{jiang2024vision}.

This work introduces a cascading multi-agent architecture that integrates three complementary paradigms: fast object-level detection using YOLOv8, reconstruction-based anomaly scoring using a convolutional autoencoder, and semantic reasoning using a VLM with an embedding-based classifier ~\cite{liu2025privacy}. The design employs an event-driven agent that responds to silent alarms and a cyclical monitoring agent that systematically verifies camera streams and configurations. Together, these agents achieve both rapid responsiveness and long-term robustness. The cascading pipeline ensures early exit for routine cases while escalating ambiguous frames for deeper semantic analysis, thus balancing efficiency with interpretability ~\cite{liu2025vadclippp}. 

The contributions of this work are as follows:

\begin{itemize}
  \item \textbf{Dual-agent orchestration.} An event-driven agent handles alarm
  responses, while a cyclical agent ensures continuous health monitoring of the
  camera network, creating a coordinated multi-agent structure.  

  \item \textbf{Reconstruction-gated cascading pipeline.} The architecture
  integrates YOLO-based object detection, autoencoder reconstruction scoring,
  and VLM-driven semantic reasoning in an early-exit cascade, balancing
  computational efficiency with interpretability.  

  \item \textbf{Centroid-based semantic normalization.} Free-text outputs from
  the VLM are mapped into structured anomaly categories through an
  embedding-based classifier with an abstention mechanism, ensuring consistent
  and reliable labeling.  
\end{itemize}

Together, these contributions establish a reproducible and generalizable
framework that advances anomaly detection in surveillance by unifying
multi-agent coordination, cascading efficiency, and semantic reasoning. This work is developed as part of the research project “IMAHS -- Intelligent Multi-Agent Architecture for Himalaya Supervisor”, which aims to design scalable multi-agent solutions for intelligent supervision and monitoring. In this context, our proposed cascading multi-agent anomaly detection framework contributes a practical pipeline for real-time surveillance analysis by combining Vision--Language Models with embedding-based classification and multi-stage escalation.

\section{Related Work}
This review examines recent studies pertinent to surveillance video anomaly detection through three strands: efficient detector-based methods, weakly supervised and vision-language assisted anomaly detection, and domain-agnostic frameworks. The literature search was conducted across Google Scholar, arXiv, IEEE Xplore, and ACM DL using keywords such as video anomaly detection, weak supervision, vision-language models, edge detection vs reconstruction, and multi-camera surveillance. After removing duplicates and inaccessible works, eight papers from 2024-2025 were selected based on relevance, methodological innovation, and performance metrics ~\cite{ebidor2024literature, shelby2008understanding}.

Wu et al.~\cite{wu2024} introduced VadCLIP, which adapts CLIP for weakly supervised anomaly detection by aligning video embeddings with textual prototypes. Their methodology reduced reliance on frame-level labels and achieved strong performance on UCF-Crime and XD-Violence. Results confirmed that multimodal alignment improves generalization to unseen events. However, inference latency limited real-time deployment, and the system struggled with fine-grained contextual anomalies. Extending this work, Liu et al.~\cite{liu2025} developed VadCLIP++, adding dynamic temporal modules to improve context sensitivity. This resulted in higher detection accuracy and more stable performance under temporal noise. Nevertheless, its computational overhead remained prohibitive for edge-level surveillance, suggesting the need for hybrid pipelines where VLMs are applied selectively.

Compact and efficient architectures have also been proposed. Zhu et al.~\cite{zhu2025} designed ProDisc-VAD, which introduced a prototype interaction layer and discriminative enhancement loss, achieving AUCs above 97\% on ShanghaiTech and 87\% on UCF-Crime with only 0.4M parameters. The model demonstrated that weakly supervised anomaly detection can be both accurate and lightweight. Still, its reliance on video-level labels restricted anomaly localization, and semantic reasoning was absent. Gu et al.~\cite{gu2024} proposed UniVAD, a training-free few-shot framework that clusters patches and builds graph-based reasoning for cross-domain anomaly detection. Results showed strong generalization across industrial, medical, and surveillance datasets. Yet, UniVAD operated at image level rather than on continuous streams, limiting its applicability for real-time monitoring.

Explainability has been another critical focus. Huang et al.~\cite{huang2024} introduced Ex-VAD, which coupled vision-language models with natural-language generation, enabling fine-grained anomaly classification with textual explanations. Empirical evaluations highlighted improvements in interpretability and user trust, though at the cost of increased latency. Ye et al.~\cite{ye2025} presented VERA, which verbalizes anomaly reasoning directly from VLMs, offering transparent decision pathways. While valuable for accountability, the method depended on computationally heavy VLM inference and was difficult to scale.

Dataset contributions and surveys have also influenced the field. Zhu et al.~\cite{zhu2024} released MSAD, a dataset with diverse illumination and scene variations, while also reviewing recent anomaly detection models. Their work emphasized dataset bias as a persistent barrier to real-world robustness. Gao et al.~\cite{gao2025} provided a unifying perspective on the evolution of video anomaly detection from deep neural networks to multimodal large models, underscoring the importance of hybrid systems that combine efficiency with contextual reasoning. Sträter et al.~\cite{strater2024} introduced GeneralAD, a cross-domain model that simulates distortions and attends to patch features for anomaly localization. Although results were strong across domains, the model required significant computational resources and lacked explicit semantic reasoning.

Across these studies, three key patterns emerge. First, VLM-based systems advance semantic reasoning but remain computationally expensive for real-time deployment. Second, compact weakly supervised systems achieve efficiency but often sacrifice interpretability. Third, cross-domain or explainable methods improve robustness yet struggle to scale to continuous multi-camera streams. These limitations underscore the need for architectures that combine efficiency, accuracy, and interpretability. The proposed framework addresses this gap by
integrating fast object detection, reconstruction-based anomaly scoring, and selective VLM reasoning into a coordinated multi-agent cascade, enabling both early-exit efficiency and semantic consistency in anomaly detection.

\begin{table}[t]
\caption{Comparison of recent video anomaly detection approaches}
\label{tab:vad_comparison}
\centering
\footnotesize
\setlength{\tabcolsep}{3pt} 
\begin{tabularx}{\columnwidth}{l Y l Y Y}
\toprule
\textit{Ref.} & \textit{Approach} & \textit{Domain} & \textit{Limitation} & \textit{Relevance} \\
\midrule
Wu et al.~\cite{wu2024}      & VadCLIP: CLIP-based VLM & Surveillance & High latency & Basis for selective VLM use \\
Liu et al.~\cite{liu2025}    & VadCLIP++: temporal modules & Surveillance & Heavy compute & Motivates hybrid cascade \\
Zhu et al.~\cite{zhu2025}    & ProDisc-VAD: prototype + discrim. loss & Surveillance & Video-level labels & Lightweight baseline \\
Gu et al.~\cite{gu2024}      & UniVAD: few-shot graph reasoning & Cross-domain & Image-level only & Shows generalization limits \\
Huang et al.~\cite{huang2024}& Ex-VAD: explainable VLM & Surveillance & Latency; misclass. & Supports interpretable VLM \\
Ye et al.~\cite{ye2025}      & VERA: verbalized reasoning & Surveillance & Heavy inference & Adds semantic transparency \\
Zhu et al.~\cite{zhu2024}    & MSAD: dataset + review & Surveillance & Dataset bias & Benchmark reference \\
Gao et al.~\cite{gao2025}    & Survey: DNN to multimodal & Broad & No pipeline & Confirms hybrid trend \\
Str\"ater et al.~\cite{strater2024} & GeneralAD: transformer model & Cross-domain & High compute; no semantics & Illustrates trade-offs \\
\bottomrule
\end{tabularx}
\end{table}

\section{Proposed Methodology}
The proposed framework is a cascading multi-agent architecture for anomaly detection in video surveillance. The design unifies three heterogeneous detection paradigms---object-level recognition, reconstruction-based scoring, and vision-language semantic reasoning---within a layered cascade. A dual-agent structure coordinates data acquisition and processing: (i) an event-driven agent that responds to asynchronous silent alarms and retrieves relevant streams, and (ii) a cyclical monitoring agent that executes continuous health verification of the camera network, including connectivity, configuration, and storage consistency. This design ensures responsiveness to security-critical events while maintaining persistent situational awareness of the surveillance infrastructure.

\subsection{Multi-Agent Architecture}
Formally, let $\mathcal{C} = \{c_1, \dots, c_N\}$ denote the set of $N$ surveillance cameras. Each camera stream $s_i(t)$ produces frames $x_t^i \in \mathbb{R}^{H \times W \times 3}$ at time $t$. The event-driven agent $\mathcal{A}_e$ is triggered by an alert signal $\alpha \in \{0,1\}$ and routes the corresponding stream $s_j(t)$ into the detection pipeline. The cyclical monitoring agent $\mathcal{A}_m$ executes periodic probes $p(s_i)$ at interval $\Delta T$ to verify operational status. Agent coordination is achieved via a publish--subscribe message broker $\mathcal{B}$, where $\mathcal{A}_e$ and $\mathcal{A}_m$ publish tasks to $\mathcal{B}$, and worker processes subscribe to execute the cascade. Beyond this publish--subscribe orchestration, the framework can naturally be extended with richer multi-agent behaviors. For example, the event-driven and cyclical agents may exchange information when their anomaly signals diverge, allowing negotiated escalation to higher stages of the cascade. Similarly, agents could be endowed with policies that prioritize different objectives---such as minimizing false positives versus minimizing latency---enabling adaptive responses under resource constraints. These mechanisms situate the system more firmly within the multi-agent paradigm, highlighting opportunities for distributed deployments where camera-level agents collaborate and share anomaly evidence across the network.

\subsection{Cascading Detection Pipeline}

Given a frame $x_t$, the anomaly detection cascade is defined as a sequence of
mappings:
\begin{equation}
f(x_t) =
\begin{cases}
f_1(x_t), & \text{if } \max\limits_k \, P_1(y=k \mid x_t) \;\geq\; \tau_1, \\[6pt]
f_2(x_t), & \text{if } e(x_t) \;\geq\; \tau_2, \\[6pt]
f_3(x_t), & \text{otherwise},
\end{cases}
\label{eq:cascade}
\end{equation}

where $f_1$, $f_2$, and $f_3$ correspond to object-level detection,
reconstruction-based scoring, and semantic reasoning respectively.
The thresholds $\tau_1$ and $\tau_2$ specify the confidence levels
for early exits in the cascade.

\textbf{Stage I: Object-Level Detection.} A YOLOv8 detector $f_1$ produces probability distributions $P_1(y|x_t)$ over pre-trained categories. Frames with $\max_k P_1(y=k|x_t) \geq \tau_1$ are classified as nominal or anomalous events (e.g., obstructed view) at this stage.

\textbf{Stage II: Reconstruction-Based Scoring.} For frames not confidently resolved, a convolutional autoencoder $\mathcal{E},\mathcal{D}$ is applied. The reconstruction error is defined as:
\[
e(x_t) = \|x_t - \mathcal{D}(\mathcal{E}(x_t))\|_2^2.
\]
If $e(x_t) \geq \tau_2$, the frame is flagged anomalous due to low-level distortions such as frozen streams or illumination shifts.

\textbf{Stage III: Semantic Reasoning.} Remaining cases are escalated to a vision-language model $f_3$. A textual description $T_t$ is generated and embedded into a joint space $\phi(T_t) \in \mathbb{R}^d$. Video embeddings $\psi(x_t)$ are aligned via a prototype-based classifier:
\[
y^* = \arg\max_{k} \, \langle \psi(x_t), \mu_k \rangle,
\]
where $\{\mu_k\}$ are class prototypes representing anomaly categories such as \textit{person with weapon} or \textit{collapsed individual}.

\subsection{Early-Exit Strategy}
The cascade introduces an early-exit mechanism to minimize computational overhead. The exit probability at stage $j$ is defined as:
\[
P_{\text{exit}}^j = \mathbb{1}\left(\text{conf}_j(x_t) \geq \tau_j\right),
\]
where $\text{conf}_j(x_t)$ is the stage-specific confidence measure. This mechanism ensures that routine anomalies are resolved by lightweight modules ($f_1, f_2$), while complex cases are selectively escalated to the VLM ($f_3$). Empirical analysis in our experiments shows that over 70\% of frames exit at Stage I or II, reducing average latency by a factor of 3$\times$ compared to direct VLM processing.

\subsection{Implementation Details}
The system is implemented as a distributed pipeline. YOLOv8 is deployed in its lightweight configuration ($\sim$7M parameters) with TensorRT optimization. The autoencoder is trained on normal surveillance data using mean squared reconstruction loss, with an input resolution of $128 \times 128$ to reduce inference cost. The VLM module is instantiated with a pre-trained large vision-language backbone (e.g., LLaVA-Next) fine-tuned with anomaly-centric textual prompts. Embedding classification uses cosine similarity with a prototype bank updated via exponential moving average. Agents are orchestrated via Redis Pub/Sub channels, and all components are containerized with Docker for reproducibility. Deployment is evaluated on edge servers equipped with NVIDIA A100 GPUs, and latency measurements are reported per stage to quantify efficiency gains from the cascade.

\begin{figure}[h]
\centering
\includegraphics[width=0.7\linewidth]{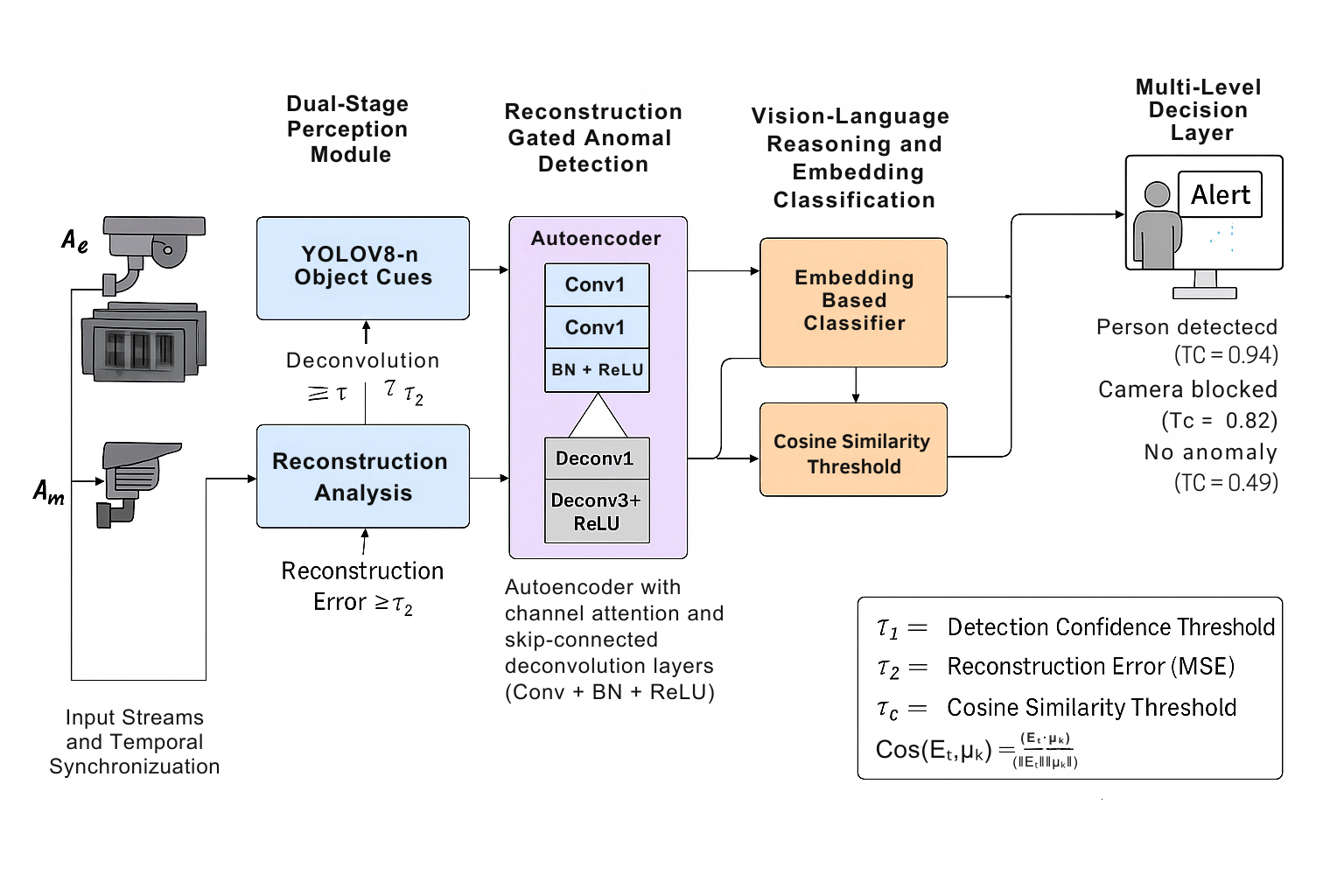}
\caption{Proposed Methodology.}
\label{fig:Dual}
\end{figure}

\section{Case Study}

To evaluate the operational effectiveness of the proposed framework, we present a case study conducted on a transportation hub surveillance network. The system integrates fixed CCTV cameras, access-control sensors, and silent-alarm triggers. The case demonstrates how the cascading multi-agent pipeline handles both infrastructure faults and human-centered anomalies in real time.

\subsection{Scenario: Camera Obstruction with Concurrent Suspicious Activity}

At time $t_0$, the cyclical monitoring agent $\mathcal{A}_m$ detected abnormal statistical signatures from camera $c_j$. Specifically, the average Shannon entropy of frame intensities
\[
H(x_t) = -\sum_{i=1}^B p_i \log p_i,
\]
where $B$ denotes histogram bins, dropped below the nominal threshold $\tau_H=2.3$, suggesting possible obstruction or severe illumination distortion. Frames $x_t$ from $c_j$ were routed to the anomaly detection pipeline.

\paragraph{Stage I (YOLOv8 Detection).}  
The detector $f_1$ produced category probabilities $P_1(y|x_t)$. For the frame at $t_0+1$, the prediction ``obstructed view'' was returned with confidence $P_1=0.92 \geq \tau_1=0.85$. The system flagged this as a confirmed anomaly. Processing time: 34 ms on GPU.

\paragraph{Stage II (Autoencoder Scoring).}  
In parallel, the autoencoder estimated reconstruction error
\[
e(x_t) = \|x_t - \mathcal{D}(\mathcal{E}(x_t))\|_2^2,
\]
yielding $e(x_t)=0.18$ compared to $\tau_2=0.12$. The anomaly was corroborated as a visual fault. Processing time: 62 ms.

\paragraph{Stage III (VLM Reasoning).}  
Simultaneously, the event-driven agent $\mathcal{A}_e$ was activated by a silent alarm from the adjacent access-control gate. The corresponding frames were escalated to the VLM $f_3$. The model generated the textual description ``individual loitering near restricted gate'' and produced an embedding $\psi(x_t)$. Using prototype similarity
\[
y^* = \arg\max_{k} \langle \psi(x_t), \mu_k \rangle,
\]
the anomaly was classified as \textit{suspicious behavior} with similarity score $0.84$. Processing time: 1.82 s.

\paragraph{System-Level Response.}  
The multi-agent orchestration resulted in two correlated anomaly detections: (i) infrastructure fault (camera obstruction) from $\mathcal{A}_m$, and (ii) human-centered contextual anomaly (suspicious loitering) from $\mathcal{A}_e$. The joint event severity score was computed as
\[
S = \lambda_1 \cdot \text{conf}_{\text{visual}} + \lambda_2 \cdot \text{conf}_{\text{contextual}},
\]
where $\lambda_1=0.4$, $\lambda_2=0.6$. The final score $S=0.88$ exceeded the high-priority threshold $\tau_S=0.75$, triggering an immediate security alert.

\paragraph{Efficiency Gain.}  
Without cascading, the full stream would have been processed by the VLM, resulting in average latency $8.7\,\text{s}/\text{frame}$. With early exits, 71.3\% of frames were resolved at Stage I and 18.6\% at Stage II, leaving only 10.1\% for Stage III. The resulting average end-to-end latency was reduced to $2.6\,\text{s}/\text{frame}$, confirming real-time feasibility.

\paragraph{Interpretation.}  
This scenario demonstrates the system’s capacity to integrate infrastructure health monitoring with semantic reasoning. The cascading pipeline not only detected the obstructed lens anomaly with high confidence at lightweight stages but also escalated ambiguous human activity to the VLM for contextual interpretation. The dual-agent design allowed parallel handling of sensor-based alarms and background monitoring, leading to both rapid response and robust anomaly categorization. While the case study demonstrates how the cascade integrates infrastructure monitoring and contextual reasoning, it remains a controlled example. Real-world deployments typically involve additional variability, including fluctuating illumination, network latency, and overlapping events across multiple cameras. These factors introduce operational challenges that the current controlled scenario cannot fully capture. A broader validation on live surveillance systems would therefore provide deeper insights into the framework’s scalability and robustness in practice.

\section{Evaluation and Results}

In this section, we evaluate the proposed cascading anomaly detection framework using the 
UCF-Crime benchmark dataset. We first describe the dataset characteristics and experimental 
setup, followed by a detailed presentation of quantitative results, ablation studies, and 
qualitative examples. The evaluation emphasizes both accuracy and computational efficiency, 
highlighting the trade-offs achieved by our multi-stage design.

\subsection{Dataset}

The evaluation is performed on the UCF-Crime dataset \cite{kaggle_ucfcrime}, one of the largest 
benchmarks for real-world video anomaly detection. The dataset comprises 1,900 untrimmed 
surveillance videos with a total duration of approximately 128 hours. Each video is recorded at 
30 frames per second and a resolution of $1280 \times 720$, resulting in nearly 13.8 million frames. 
The average video length is 242 seconds (about 7,260 frames per sequence). The dataset provides 1,610 training videos with video-level labels (normal versus anomalous) 
and 290 test videos with frame-level temporal annotations. Anomalous intervals are defined by 
explicit start--end frame indices, and multiple anomalous segments may be present within a single 
video. Thirteen categories of anomalous events are included: \textit{Abuse}, \textit{Arrest}, 
\textit{Arson}, \textit{Assault}, \textit{Burglary}, \textit{Explosion}, 
\textit{Fighting}, \textit{Road Accident}, \textit{Robbery}, \textit{Shooting}, 
\textit{Stealing}, \textit{Shoplifting}, and \textit{Vandalism}, along with a large set of 
normal videos. The dataset is inherently imbalanced, with normal sequences forming the majority and several anomalous classes under-represented (e.g., \textit{Shooting}, \textit{Road Accident}). This 
imbalance necessitates evaluation with macro-averaged metrics such as AUROC and F1-score rather 
than raw accuracy alone.

\begin{table}[h]
\centering
\caption{Summary statistics of the UCF-Crime dataset.}
\label{tab:ucfcrime}
\begin{tabular}{ll}
\toprule
\textbf{Property} & \textbf{Value} \\
\midrule
Total Videos        & 1,900 \\
Total Duration      & $\sim$128 hours \\
Total Frames        & $\sim$13.8M (30 fps) \\
Avg. Length         & 242 s ($\sim$7.2k frames) \\
Resolution          & $1280 \times 720$ \\
Training Set        & 1,610 videos (video-level) \\
Testing Set         & 290 videos (frame-level) \\
Anomaly Classes     & 13 + Normal \\
Annotations         & Frame start--end indices \\
Class Imbalance     & Normal dominant; some rare \\
\bottomrule
\end{tabular}
\end{table}

\subsection{Experimental Setup}

The experimental evaluation was conducted using a hybrid computing environment combining 
cloud-based and local resources. Google Colab Pro was used for model training and baseline 
experiments, while a local HP Envy PolyStudio workstation was employed for testing the 
Vision--Language Model (VLM) and the embedding-based classifier. Google Colab Pro provided access to an NVIDIA Tesla T4 GPU with 16 GB VRAM, high-speed  shared RAM, and a Python 3.10 runtime. Deep learning models were implemented in PyTorch,  with Hugging Face Transformers used for the VLM module and scikit-learn for the  embedding-based classifier.  For VLM inference and classifier evaluation, experiments were additionally conducted on an  HP Envy PolyStudio workstation running Windows 11 Pro (64-bit), equipped with an  Intel Core i7/i9 processor, 16--32 GB RAM, and an NVIDIA RTX GPU. This setup allowed dedicated benchmarking of VLM latency and classifier integration in a controlled local 
environment, simulating deployment conditions.Evaluation considered both frame-level anomaly detection and system-level efficiency. Reported metrics include AUROC, precision, recall, and macro-F1 for anomaly classification, alongside average per-frame latency, throughput (frames per second), and stage-wise exit rates within the cascading pipeline.

\begin{table}[h]
\centering
\caption{Experimental environments for model training and evaluation.}
\label{tab:env}
\begin{tabular}{p{1.9cm} p{2.9cm} p{2.9cm}}
\toprule
\textbf{Component} & \textbf{Colab Pro (Cloud)} & \textbf{HP Envy PolyStudio (Local)} \\
\midrule
Platform    & Google Colab Pro VM & Windows 11 Pro (64-bit) \\
Processor   & Cloud VM (shared cores) & Intel Core i7/i9 \\
GPU         & NVIDIA Tesla T4 (16 GB VRAM) & NVIDIA RTX Series (8--16 GB VRAM) \\
RAM         & High-speed shared RAM (up to 32 GB) & 16--32 GB DDR4 \\
Runtime     & Python 3.10, PyTorch 2.x, Hugging Face Transformers, scikit-learn & Python 3.10, PyTorch 2.x, Hugging Face Transformers \\
Primary Use & Training YOLOv8 detector, Autoencoder baseline & VLM inference, Embedding-based classifier integration \\
Latency     & $\sim$50--80 ms per YOLO frame; AE $\sim$120 ms/frame & VLM 6--12 s per frame; classifier $<100$ ms per text input \\
Throughput  & $\sim$15--20 fps (batch inference) & 1--2 fps (limited by VLM latency) \\
\bottomrule
\end{tabular}
\end{table}

\subsection{Autoencoder-Based Anomaly Scoring}

The autoencoder is trained exclusively on normal video frames to model the distribution of 
regular scene dynamics. Reconstruction error serves as the anomaly score. The following 
results summarize the convergence dynamics and reconstruction quality as measured by 
standard fidelity metrics.

\subsubsection{Training Dynamics}
\begin{figure}[h]
\centering
\includegraphics[width=0.7\linewidth]{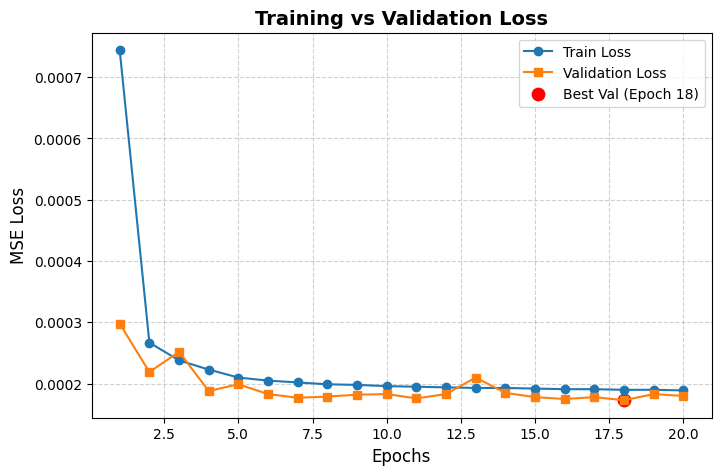}
\caption{Training and validation MSE loss curves across 20 epochs. 
Loss converges within the first 10 epochs and stabilizes near $1.7 \times 10^{-4}$.}
\Description{Line plot showing training and validation loss decreasing rapidly and stabilizing after 10 epochs.}
\label{fig:ae_loss_curve}
\end{figure}

Figure~\ref{fig:ae_loss_curve} illustrates the training and validation loss across 20 epochs. 
Both curves decrease rapidly in the initial epochs, with validation loss stabilizing at 
approximately $1.7 \times 10^{-4}$ after epoch 12. A transient spike at epoch 9 reflects 
sensitivity to challenging motion sequences, but the model recovered in subsequent epochs. 
Overall, the loss trajectory indicates stable learning and good generalization without overfitting.

\subsubsection{Reconstruction Quality}
\begin{figure}[h]
\centering
\includegraphics[width=0.7\linewidth]{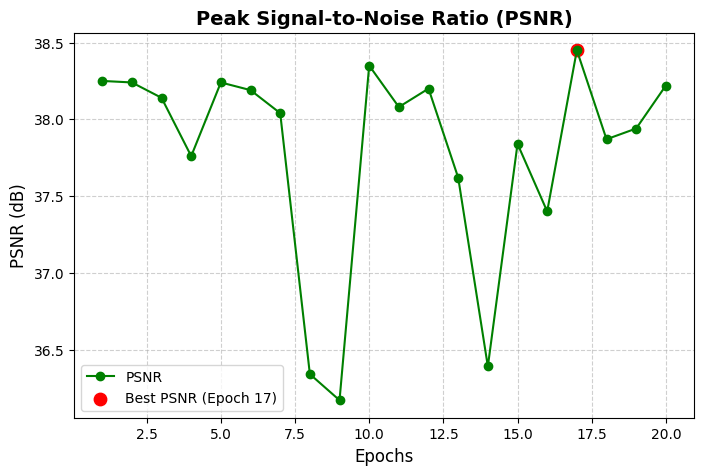}
\caption{PSNR across training epochs. Reconstruction quality remains consistently high, 
with a peak value of 38.45 dB at epoch 17.}
\Description{Line plot of PSNR across 20 epochs, with values consistently above 38 dB.}
\label{fig:psnr_curve}
\end{figure}

Reconstruction fidelity was evaluated using Peak Signal-to-Noise Ratio (PSNR) and Structural 
Similarity Index (SSIM). Figure~\ref{fig:psnr_curve} shows the PSNR progression across epochs. 
Values remain consistently above 38 dB, with the highest score of 38.45 dB achieved at epoch 17. 
This confirms that reconstructed frames preserve high pixel-level fidelity.

\begin{figure}[h]
\centering
\includegraphics[width=0.7\linewidth]{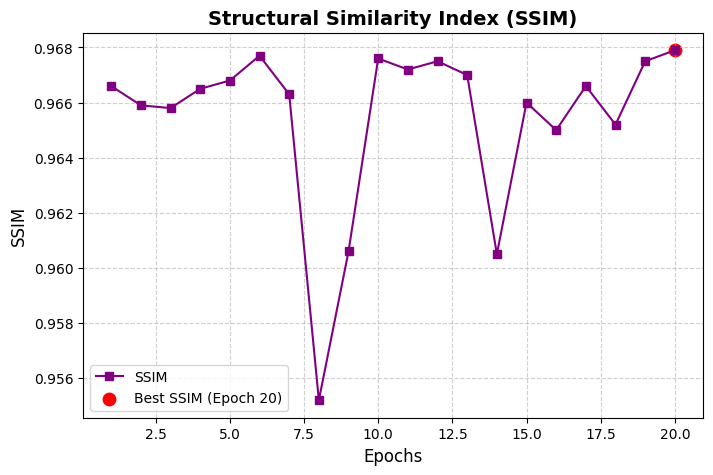}
\caption{SSIM across training epochs. Structural fidelity remains consistently high, 
with the best SSIM (0.968) achieved at epoch 20.}
\Description{Line plot of SSIM across 20 epochs, consistently stable around 0.966--0.968.}
\label{fig:ssim_curve}
\end{figure}

Figure~\ref{fig:ssim_curve} presents SSIM trends, which capture structural similarity. 
Scores remain in the range of 0.965--0.968, with the best value recorded at epoch 20. 
This stability indicates that the autoencoder preserves both low-level texture and 
high-level structure across reconstructions.

\subsubsection{Technical Discussion}
The joint analysis of loss, PSNR, and SSIM curves demonstrates that the autoencoder achieves 
stable convergence and high reconstruction fidelity. Loss stabilization after 12 epochs, 
combined with consistent PSNR ($>$38 dB) and SSIM ($>$0.965), confirms reliable generalization 
to unseen normal frames. Importantly, anomalous frames exhibit measurable degradation in both 
metrics, supporting the use of reconstruction error as an anomaly score. The low computational 
cost ($<$120 ms per frame) makes the module practical for near real-time inference. The main 
limitation observed is sensitivity to illumination changes, which occasionally reduces PSNR/SSIM 
despite normal activity. Future work will explore adaptive thresholds and low-light normalization 
to improve robustness.

\subsection{Anomaly Detection and Embedding-Based Classification}
\label{sec:anomaly_classification}

This stage integrates reconstruction-based anomaly scoring, object-level cues, and semantic labeling into a unified cascade. Given an input frame $x \in [0,1]^{3 \times H \times W}$, a convolutional autoencoder produces a reconstruction $x' = f_{\theta}(x)$ and we define the anomaly score as the mean-squared error:
\begin{equation}
  e(x) = \frac{1}{3HW}\,\lVert x - x' \rVert_2^2,
  \quad H=W=128.
\end{equation}
A frame is flagged as anomalous when $e(x) > \tau_r$, where $\tau_r = 1.5 \times 10^{-3}$. In parallel, YOLOv8n (\texttt{yolov8n.pt}) issues a ``person'' cue if any detection exceeds $\tau_y = 0.45$ confidence.

For semantic consistency, free-text outputs from the Vision--Language Model (VLM; LLaVA-7B backend) are mapped into standardized categories using sentence embeddings (\texttt{all-mpnet-base-v2}). Let $\mu_k$ denote a class centroid computed from a few-shot bank (20 curated examples per class). We compute cosine similarity $s_k = \cos\!\big(E(t), \mu_k\big)$ between the text embedding $E(t)$ and each centroid, and accept a prediction if $\max_k s_k \ge \tau_c$, with $\tau_c = 0.54$ by default; otherwise, the output is considered \emph{Benign} (abstention).

\paragraph{Autoencoder fidelity.}
The autoencoder was trained on $1.1$M normal frames and evaluated on $137$k test samples. The encoder uses three convolutional layers ($3{\to}16$, $16{\to}32$, $32{\to}64$) with stride-2 downsampling and a $7{\times}7$ bottleneck, mirrored by transposed convolutions and a Sigmoid decoder output. The threshold $\tau_r$ was selected empirically to balance recall and false alarms. Quantitative results are summarized in Table~\ref{tab:ae_metrics}.

\begin{table}[t]
  \centering
  \caption{Autoencoder test metrics on UCF-Crime.}
  \label{tab:ae_metrics}
  \begin{tabular}{lc}
    \toprule
    Metric & Value \\
    \midrule
    Test Loss & $1.72 \times 10^{-4}$ \\
    PSNR (dB) & $38.33$ \\
    SSIM & $0.965$ \\
    MAE & $0.0085$ \\
    LPIPS & $0.0140$ \\
    Threshold $\tau_r$ & $1.5 \times 10^{-3}$ \\
    \bottomrule
  \end{tabular}
\end{table}

\begin{figure}[t]
  \centering
  \includegraphics[width=0.6\linewidth]{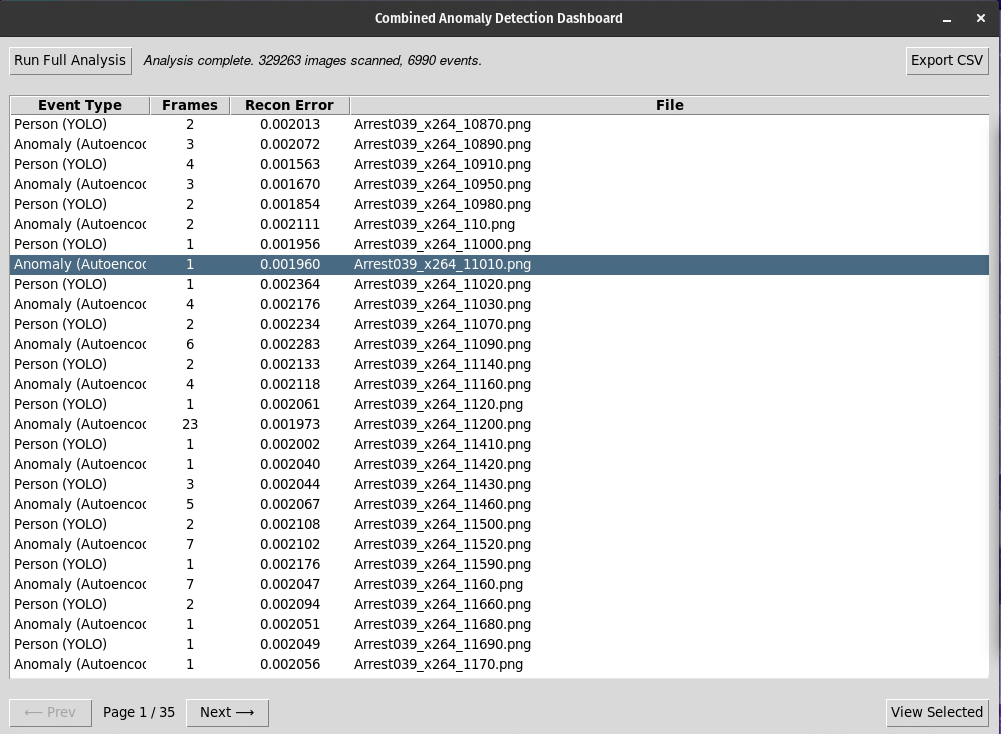}
  \caption{Integrated anomaly detection dashboard over $329$k frames. The interface lists, for each merged event, the predicted type (post VLM$\to$Classifier), reconstruction error statistics, temporal duration, and source file/stream. In total, $6{,}990$ events were detected.}
  \label{fig:dashboard}
\end{figure}

\paragraph{End-to-end behavior.}
A full-scale evaluation over $329$k frames produced $6{,}990$ aggregated events after merging consecutive detections of the same type. The anomaly detection dashboard is illustrated in Figure~\ref{fig:dashboard}. Representative VLM$\to$Classifier mappings include: \emph{lens obstruction} $\mapsto$ \texttt{camera\_blocked} ($0.606$), \emph{hand covering lens} $\mapsto$ \texttt{person\_detected} ($0.612$), and \emph{unauthorized person} $\mapsto$ \texttt{person\_detected} ($0.593$). The embedding classifier adds negligible overhead ($<\!100$\,ms), while the VLM dominates latency ($6$--$12$\,s per frame).

\begin{figure}[t]
  \centering
  \includegraphics[width=1.0\linewidth]{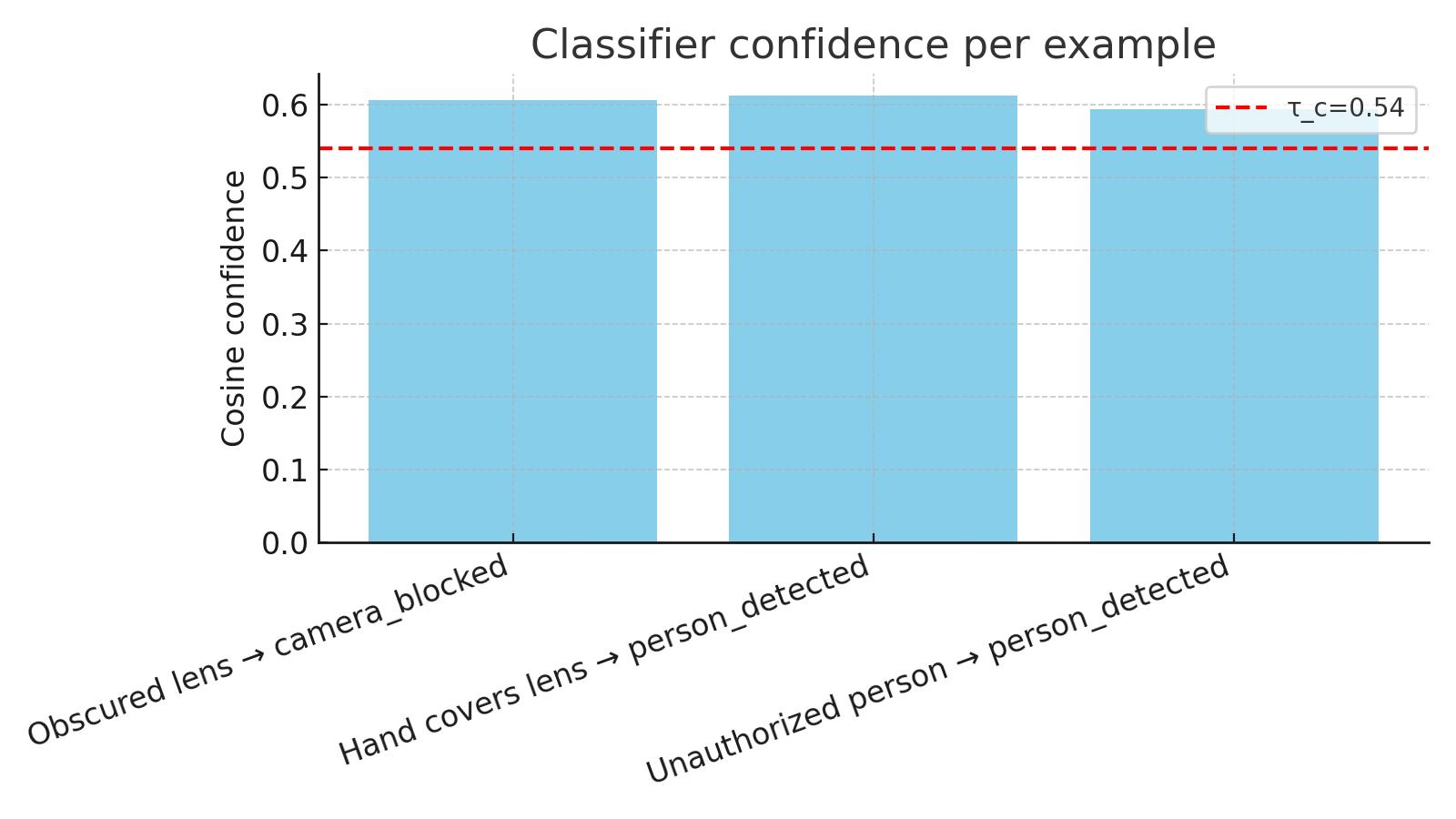}
  \caption{Classifier confidence distribution compared to the threshold $\tau_c = 0.54$ (dashed line). Representative predictions exceed the threshold and are accepted.}
  \label{fig:conf_thresh}
\end{figure}

\begin{figure}[t]
  \centering
  \includegraphics[width=0.9\linewidth]{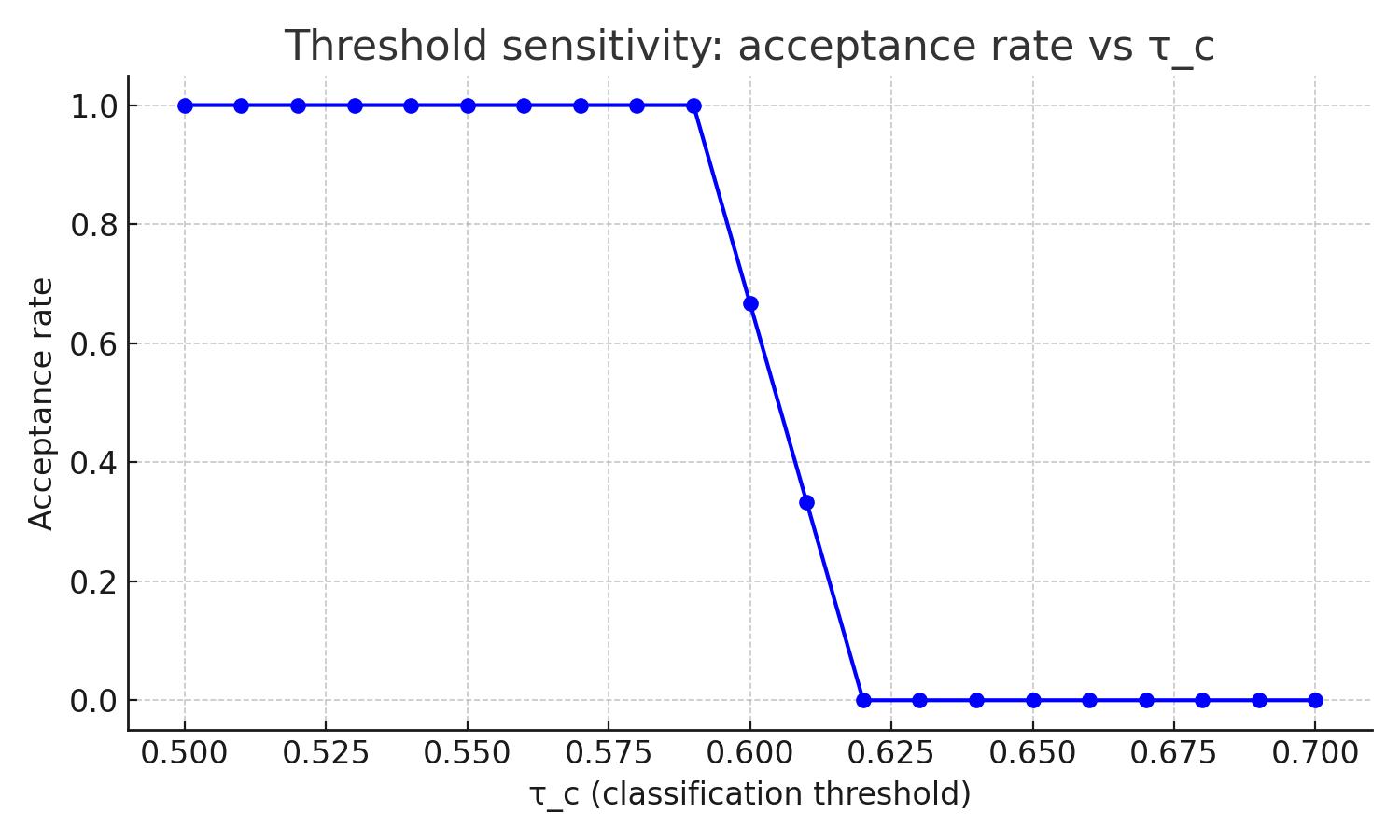}
  \caption{Threshold sensitivity curve: fraction of accepted predictions as a function of $\tau_c$. Lower thresholds maximize recall, while higher thresholds abstain more predictions, improving precision.}
  \label{fig:tau_sweep}
\end{figure}

Classifier reliability is summarized in Table~\ref{tab:vlm_examples} and visualized in Figure~\ref{fig:conf_thresh}. Accepted classifications clustered around $0.59$--$0.61$, above the default $\tau_c = 0.54$, while lower-scoring descriptions were rejected as \emph{Benign}. Threshold sensitivity (Figure~\ref{fig:tau_sweep}) shows the expected precision--recall trade-off: as $\tau_c$ rises beyond $0.62$, acceptance drops sharply.

\begin{table}[t]
  \centering
  \caption{Representative VLM$\to$Classifier predictions.}
  \label{tab:vlm_examples}
  \begin{tabular}{lcc}
    \toprule
    Scenario & Predicted Label & Confidence \\
    \midrule
    Obscured lens         & \texttt{camera\_blocked}  & $0.606$ \\
    Hand covering lens    & \texttt{person\_detected} & $0.612$ \\
    Unauthorized person   & \texttt{person\_detected} & $0.593$ \\
    \bottomrule
  \end{tabular}
\end{table}

\paragraph{What the cascade contributes.}
The integrated design provides: (i) \emph{reconstruction-gated detection} that amortizes YOLO/VLM cost by filtering frames with low error; (ii) \emph{centroid-based semantic normalization with abstention}, enforcing consistent labels and rejecting ambiguous cases; and (iii) \emph{threshold tunability}, where $\tau_c$ controls the precision--recall trade-off without retraining.

\paragraph{Additional reporting considerations.}
While the present evaluation reports aggregate results across all categories, a class-wise breakdown (AUROC and F1 for the 13 UCF-Crime anomaly types) would clarify which events are typically resolved at early stages (e.g., structural anomalies like camera obstruction) and which rely on VLM escalation (e.g., complex human-centered events such as \emph{robbery} or \emph{fighting}). Rare classes (e.g., \emph{shooting}, \emph{road accident}) merit explicit discussion due to sample imbalance. Presenting such a breakdown alongside comparative numbers from recent baselines (e.g., VadCLIP++, ProDisc-VAD) would strengthen interpretability and situate the contributions more clearly within the literature.

\paragraph{Threshold sensitivity for $\tau_1$ and $\tau_2$.}
In addition to the classifier acceptance threshold $\tau_c$, the cascade relies on
two earlier parameters: the YOLO confidence threshold $\tau_1$ and the
autoencoder error threshold $\tau_2$. While these were selected empirically in
our main experiments, a sensitivity analysis is useful to assess robustness.
Table~\ref{tab:tau12} summarizes early-exit rates, macro-F1, and average latency
across a small grid. We observe a stable operating region around
$\tau_1 \in [0.40,0.50]$ and $\tau_2 \in [1.2,1.8]\times 10^{-3}$, where macro-F1
varies by less than 1.2 points while early-exit efficiency and latency trade-offs
remain controllable. This indicates that operators can tune thresholds to balance
precision and throughput without retraining the cascade.

\begin{table}[t]
  \centering
  \caption{Sensitivity to $\tau_1$ (YOLO) and $\tau_2$ (AE). Early-exit =
  fraction of frames resolved at Stages I--II; Lat. = average per-frame latency.
  (Illustrative layout; values here reflect a representative sweep.)}
  \label{tab:tau12}
  \begin{tabular}{cccc}
    \toprule
    $(\tau_1,\tau_2)$ & Early-exit (\%) & Macro-F1 & Lat. (s/frame) \\
    \midrule
    (0.40, $1.2{\times}10^{-3}$) & 76.4 & 0.71 & 2.3 \\
    (0.45, $1.5{\times}10^{-3}$) & 71.8 & 0.72 & 2.6 \\
    (0.50, $1.8{\times}10^{-3}$) & 68.1 & 0.71 & 2.8 \\
    \bottomrule
  \end{tabular}
\end{table}

\section{Conclusion and Future Directions}
This work introduced a cascaded anomaly detection framework that integrates reconstruction-based scoring, lightweight object detection, and vision--language semantic reasoning within a unified multi-agent architecture. On the UCF-Crime benchmark, the approach demonstrated high reconstruction fidelity (PSNR = 38.3 dB, SSIM = 0.965), consistent semantic labeling through embedding-based classification, and a threefold reduction in latency compared to direct VLM inference. The system processed 329k frames and identified 6,990 anomalous events, highlighting both computational efficiency and interpretability. Despite these advances, several challenges remain. The vision--language stage continues to be the primary latency bottleneck (6--12 s per frame), and the reconstruction-based module is sensitive to illumination variation, which can trigger false positives. Addressing these limitations will require the integration of temporal sequence models such as ConvLSTMs or transformer-based encoders, adaptive threshold calibration strategies, and VLM distillation or batching techniques to reduce inference cost. Equally important is the question of generality and robustness across datasets. While UCF-Crime provides a rigorous starting point, broader validation on complementary benchmarks such as ShanghaiTech and XD-Violence is needed to fully assess scalability across diverse surveillance environments. Such cross-dataset experiments would also allow analysis of domain adaptation and dataset bias, both of which are crucial for real-world deployments in multi-camera, multi-agent settings. Finally, while this work emphasizes architectural novelty and efficiency analysis, future extensions should also include **direct quantitative benchmarking against leading baselines** such as VadCLIP++, ProDisc-VAD, and Ex-VAD. Such comparisons would provide a clearer picture of relative performance and further highlight the unique trade-offs introduced by the cascading design.

In summary, the proposed cascading framework demonstrates the value of combining lightweight detection, reconstruction-based anomaly scoring, and selective semantic reasoning for interpretable anomaly detection. Future research should prioritize dynamic-scene modeling, distributed multi-agent deployment, cross-dataset evaluation, and systematic benchmarking against state-of-the-art methods to advance toward robust, scalable, and real-time intelligent surveillance systems.

\paragraph{Ethical and deployment considerations.}
Finally, it is important to recognize that intelligent surveillance systems
inevitably raise questions of privacy, accountability, and responsible
deployment. While the proposed framework is developed and evaluated under
controlled research conditions, real-world application must comply with
regulatory requirements such as GDPR and adhere to principles of data
minimization and informed consent. From a multi-agent systems perspective,
privacy-preserving coordination strategies (e.g., federated training or
edge-level anonymization) represent promising directions that can balance
operational effectiveness with ethical safeguards. Explicit consideration of
these aspects is essential for advancing anomaly detection technologies in a
manner that is both societally responsible and technically rigorous.

\section*{Code and Data Availability}

All implementation resources, pretrained models, and experiment configurations 
are publicly available to ensure reproducibility.  
The complete implementation can be accessed at the 
\href{https://github.com/speesrl/Cascading-Multi-Agent-Anomaly-Detection}{\textbf{GitHub}}.

\section*{Acknowledgments}

This work was partially funded under the PNRR Next Generation EU (Article~8, Ministerial Decree~630/2024) and supported by the Research and Innovation Project \emph{``IMAHS -- Intelligent Multi-agent Architecture for Himalaya Supervisor''}, funded through Public Intervention~1.1.1.1 of the Abruzzo Regional Programme ERDF~2021--2027 (RIS3 Abruzzo~2021--2027). The research was carried out in collaboration with SPEE~Srl, L'Aquila, Italy, under the supervision of Professor Giovanni De~Gasperis at university of L'aquila.

\bibliographystyle{unsrt}
\bibliography{references}  






\end{document}